%% file: main.tex
\documentclass[preprint,12pt]{elsarticle}

\usepackage{amssymb}

\usepackage{framed,multirow}

\usepackage{amssymb}
\usepackage{latexsym}

\usepackage{url}
\usepackage{xcolor}
\usepackage{hyperref}
\usepackage{booktabs}
\usepackage{multirow}
\usepackage{makecell}
\usepackage[nooneline]{caption} 
\usepackage{stfloats}
\usepackage{tabularx}
\usepackage{colortbl}
\usepackage{filecontents}
\usepackage{amsmath,amssymb,amsfonts}
\usepackage{algorithmic}
\usepackage{graphicx}
\usepackage{textcomp}
\usepackage{soul}
\usepackage{algorithm}
\usepackage{algorithmic}
\usepackage{xcolor}
\usepackage{wrapfig}
\usepackage{ifthen}
\usepackage{marvosym}
\definecolor{mypink}{rgb}{.99,.91,.95}
\definecolor{mygreen}{RGB}{107,147,147}
\definecolor{lgreen}{HTML}{00b8a9}
\definecolor{mygray}{HTML}{eeeeee}

\newcommand{\tabincell}[2]{\begin{tabular}{@{}#1@{}}#2\end{tabular}}

\newif\ifisClear
\isCleartrue

\newcommand{\textred}[1]{
	\ifisClear
    	{#1}
  	\else
    	\textcolor{red}{#1}
  	\fi 
}

\journal{Neurocomputing}

\begin{document}

\begin{frontmatter}

\title{Class Attention to Regions of Lesion for Imbalanced Medical Image Recognition}

\author[inst1]{Jia-Xin Zhuang}
\author[inst2]{Jiabin Cai}
\author[inst3,inst4]{Jianguo Zhang}
\author[inst2]{Wei-shi Zheng}
\author[inst2,inst4]{Ruixuan Wang\Letter }

\affiliation[inst1]{
            organization={Department of Computer Science and Engineering},
            addressline={Hong Kong University of Science and Technology}, 
            state={Hong Kong},
            country={China}}
            
\affiliation[inst2]{
            organization={Department of Computer Science and Engineering},
            addressline={Sun Yat-sen University}, 
            city={Guangzhou},
            postcode={510006}, 
            state={China}}

\affiliation[inst3]{
            organization={Research Institute of Trustworthy Autonomous Systems and Department of Computer Science and Engineerin},
            addressline={Southern University of Science and Technology}, 
            country={China}}
            
\affiliation[inst4]{
            addressline={Peng Cheng Laboratory}, 
            country={China}}

\input{sec/abstract.tex}


\end{frontmatter}

\input{sec/introduction.tex}
\input{sec/relatedWork.tex}
\input{sec/method.tex}

\input{sec/experiments.tex}

\input{sec/conclusion.tex}

\input{sec/appendix.tex}

\bibliographystyle{elsarticle-num} 
\bibliography{IEEEabrv,NameFull,cas-refs,cas-refs-2}

\end{document}

%% file: sec/abstract.tex
\begin{abstract}
Automated medical image classification is the key component in intelligent diagnosis systems. However, most medical image datasets contain plenty of samples of common diseases and just a handful of rare ones, leading to major class imbalances.
Currently, it is an open problem 
in intelligent diagnosis to effectively learn from imbalanced training data. In this paper, we propose a simple yet effective framework, named \textbf{C}lass \textbf{A}ttention to \textbf{RE}gions of the lesion (CARE), to handle data imbalance issues by embedding attention into the training process of \textbf{C}onvolutional \textbf{N}eural \textbf{N}etworks (CNNs). The proposed attention module helps CNNs attend to lesion regions of rare diseases, therefore helping CNNs to learn their characteristics more effectively. In addition, this attention module works only during the training phase and does not change the architecture of the original network, so it can be directly combined with any existing CNN architecture. The CARE framework needs bounding boxes to represent the lesion regions of rare diseases. To alleviate the need for manual annotation, we further developed variants of CARE by leveraging the traditional saliency methods or a pretrained segmentation model for bounding box generation. Results show that the CARE variants with automated bounding box generation are comparable to the original CARE framework with \textit{manual} bounding box annotations.
A series of experiments on an imbalanced skin image dataset and a pneumonia dataset indicates that our method can effectively help the network focus on the lesion regions of rare diseases and remarkably improves the classification performance of rare diseases.
\end{abstract}

%% file: sec/introduction.tex
\section{Introduction}\label{sec:introduction}

\textbf{C}onvolutional \textbf{N}eural \textbf{N}etworks (CNNs) are being widely used for image classification, object detection, segmentation, registration, and many other tasks in medical image analysis~\cite{anwar2018medical,shen2017deep}.
With the help of CNNs, computer aided diagnosis systems can automatically recognize numerous diseases, such as breast malignancy classification~\cite{haarburger2019multi}, ear disease~\cite{viscaino2022color}, skin lesion classification~\cite{goceri2020comparative,gocceri2020convolutional}, and skin cancer detection~\cite{goceri2021automated} and run on different devices~\cite{goceri2021diagnosis}.  Generally, images of common diseases are easier to collect, while images of rare diseases are not, which may lead to data imbalance issues for model training~\cite{griggs2009clinical}. As shown in Figure~\ref{fig:dataset}, images from the major classes dominate the whole medical image dataset, while other classes consist of fewer images. That means a neural network is more likely to effectively learn the features of the common diseases than those of the rare diseases due to the very limited data of the latter classes, resulting in diagnosis bias towards the common diseases.
To alleviate this data imbalance issue, we need to effectively handle the data imbalance between common diseases and rare ones~\cite{goceri2020image,yoon2019generalizable}.

\input{figures/dataset.tex}

Several strategies have been proposed to address this problem, mainly focusing on how to effectively improve the classification performance of small sample classes. These strategies range from the input side of model training, such as balancing data between classes~\cite{sun2007cost,rahman2013addressing}, the model architecture, such as ensemble models~\cite{li2007classifying,wang2009diversity}, to the output side of model training, such as setting different weights to different classes of training samples~\cite{sun2007cost,domingos1999metacost}. 
 Different from these existing studies, which consider each image as the basic unit and mainly focus on reweighting images or classes, a novel method called CARE (\textbf{RE}gions of the lesion) is proposed here by delving into images and considering the high level semantics of images. Specifically, inspired by the process of human learning, attention was embedded into the learning process of CNN classifiers, particularly for rare diseases. By enabling classifiers to pay more attention to the lesion regions of minority class(es) during learning, the classifiers can learn disease characteristics more effectively from small samples of the minority classes. Due to limited training data for rare diseases, annotation of lesion regions from those images (in the form of bounding boxes containing lesion regions) does not usually take much effort for clinicians and thus is reasonably acceptable.
Alternatively, lesion regions could be automatically estimated in advance based on state of the art saliency detection techniques DSR~\cite{li2013saliency} and the segmentation model DeepLabV3~\cite{chen2017rethinking}, as demonstrated in our study.

Also, different from existing attention relevant deep learning studies where attention is estimated as intermediate outputs of neural networks~\cite{vaswani2017attention,xu2015show}, the proposed CARE framework novelly and explicitly uses attention as part of supervision signal (in addition to image labels) to help train the CNN classifiers. The proposed attention embedding mechanism does not alter neural network architectures and, therefore can be directly embedded into the training of any existing CNN architecture. What's more, the proposed CARE is independent of any existing approach to data imbalance, and therefore can be combined to handle the imbalance issue together. Comprehensive experiments on a skin image dataset and a pneumonia chest X-ray dataset and with multiple CNN architectures showed that paying attention to lesion regions of rare diseases during learning did improve the classification performance on rare diseases. 

It should be noted that a preliminary version of this work was presented at MIDL 2019~\cite{zhuang2019care}. We further extend our work mainly in two folds. Firstly, extensive experiments are conducted to demonstrate the effectiveness of our proposed method. Our method is robust to the selection of hyperparameters designed in the loss. Secondly, our proposed method relies on bounding boxes for small classes, which were previously labeled by human experts. 


Here we used two existing methods (DSR and DeepLabV3) to automatically detect bounding boxes~\cite{li2013saliency,chen2017rethinking}. With the help of these automatically generated bounding boxes for the small class, our method still achieves comparable classification performance on the imbalanced dataset. Therefore, our method can be applied on datasets with severe data imbalance without the need of manual bounding boxes, thus reducing the annotation efforts of domain experts.

%% file: figures/dataset.tex
\begin{figure}[htb]
    \centering
    \includegraphics[width=1\textwidth]{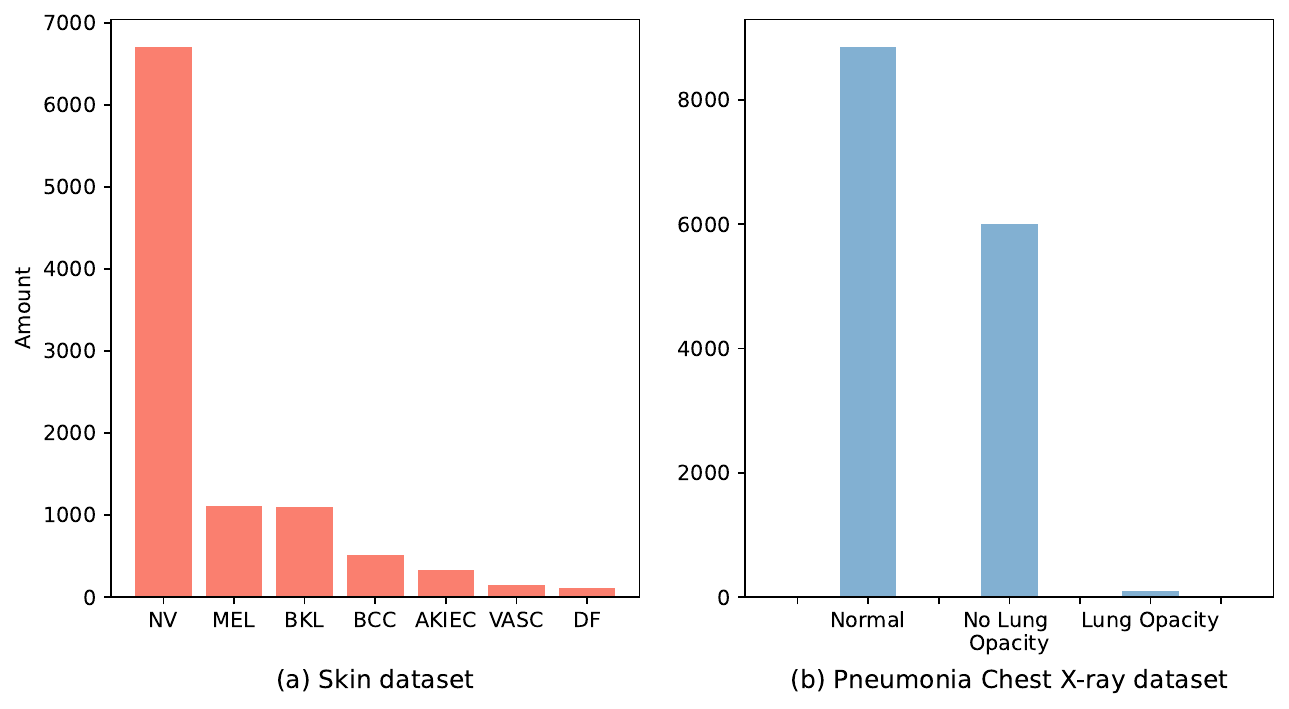}
    \caption{The imbalanced data distribution of two commonly used medical image datasets: (a) Skin dataset, and (b) Pneumonia Chest X-ray dataset. NV: Melanocytic nevus; MEL: Melanoma; BKL: Benign keratosis (solar lentigo/seborrheic keratosis/lichen planus-like keratosis); BCC: Basal cell carcinoma; AKIEC: Actinic keratosis/Bowen’s disease (intraepithelial carcinoma);  VASC: Vascular lesion; DF: Dermatofibroma.}
    \label{fig:dataset}
\end{figure}

%% file: sec/relatedWork.tex
\section{Related work}\label{sec:relatedWork}

In this section, we briefly review existing work most relevant to our proposed approach, including the approaches to data imbalance issue and existing attention mechanisms.

\subsection{Approaches to data imbalance issue}
Multiple approaches have been proposed to solve the data imbalance issue. There are mainly three mainstreams: data rebalancing, class balanced loss, and transfer learning.

On the model input side, the data rebalancing idea has been widely used. For example, one traditional approach is to over-sample the limited data available for the small sample classes~\cite{chawla2002smote} or down-sample the data of the larger sample classes~\cite{Drummond2003C4}, thus generating a similar number of training examples for each class. Data augmentation, now a default choice for training deep neural networks, can also be used as an oversampling method to generate more data for small sample classes~\cite{perez2017effectiveness}.

On the model output side, setting different weights for different training classes (class-level) or samples (sample-level) to balance loss was often used to handle the imbalance issue. At class-level reweighting training, cost sensitive learning is to improve the cost of misclassifying each training example coming from small sample classes, which can be easily realized by setting larger weights for small sample classes in the loss function~\cite{Khan2015CostSensitiveLO}. In this way, different classes as a whole were treated equally in the cost calculation. Data imbalance issues could lead to the problem of inefficiency in utilizing examples from large classes. To tackle this challenge, the normal strategy is to adaptively reweight at the sample level. Focal loss reshapes the standard cross entropy by adaptively setting lower weights to  easy samples and focusing on learning from hard samples~\cite{lin2017focal}. Similarly, reweight training assigns weight to every sample depending on their gradient direction~\cite{Ren2018LearningTR}. Based on such weights, hard negative mining can be adopted to select just a subset of training data for the next round training of classifiers~\cite{li2007classifying}. Some studies show that metric learning, such as Siamese model~\cite{hadsell2006dimensionality} and triplet loss~\cite{Balntas2016LearningLF} can also perform well with  imbalanced dataset.

Besides these approaches, another set of approaches focus on the model itself to alleviate the influence of data imbalance. Among this set, transfer learning via finetuning a pretrained classifier has been shown helpful to improve performance for both large and small sample classes~\cite{ma2019multi,wang2018chestnet}, and ensembling of multiple individual models has also become a routine to improve classification performance on each class~\cite{li2018fully}.
Different from all the existing approaches, the proposed method in this study makes use of semantic information in terms of attention particularly in images of small sample classes to solve the imbalance issue.

\subsection{Attention mechanisms}
Attention mechanisms refer to those approaches which help models focus on more task relevant information during feature extraction from original data. The basic idea is to adaptively estimate the importance of each component (e.g., image feature of each local region) and then use the weighted components for further processing. 
Originally developed for machine translation tasks~\cite{vaswani2017attention}, attention mechanisms have been recently extended and applied in computer vision~\cite{woo2018cbam}, e.g., with spatial attention in image captioning tasks~\cite{chen2017sca}, channel attention for image classification tasks~\cite{hu2018squeeze}, and self attention in video tasks~\cite{parmar2019stand}. In order to apply such attention mechanisms, the original neural network architecture would often be modified by embedding an attention module into the network model. Besides these approaches, another type of attention mechanism particularly in classification tasks, is to understand the decision process, by localizing and visualizing local image regions which contribute more to the final prediction. 
The well known methods include the \textbf{C}lass \textbf{A}ctivation \textbf{M}ap (CAM)~\cite{zhou2016learning} and its variants, such as \textbf{G}radient-weighted \textbf{C}lass \textbf{A}ctivation \textbf{M}apping (Grad-CAM)~\cite{selvaraju2017grad}. \textred{CAM and Grad-CAM can be used to generate the heatmap of visualization of a region in an image that is most relevant to a specific prediction made by a CNN. The CAM approach utilizes the Global Average Pooling layer (GAP) to combine the feature maps of the final convolution layers and subsequently generates the heatmap through a linear combination of these maps. However, this methodology may be restricted when dealing with intricate networks containing numerous linear layers that succeed the last convolution layer. In contrast, Grad-CAM addresses this limitation by leveraging gradient information, which can be effortlessly acquired through backpropagation.} However, the CAM variants were mainly developed for \textit{visualization} purpose~\cite{shinde2019hr} and, therefore, didn't seek to improve the classification or segmentation performance, since they are not directly involved in the training of neural network models.
In our study, we take a different perspective and propose an attention based framework for training the neural network, which enables the resulting network being able to handle the data imbalance problem and improve the classification performance, especially on the class of smaller size. 

%% file: sec/method.tex
\section{Method}\label{sec:method}

A study conducted in~\cite{krupinski2010current} showed that for medical students, during their training and learning process, lesions containing distinct characteristics of certain diseases were often shown and highlighted on the medical images, i.e., a kind of attention to task relevant regions.
With the help of such attention to lesion regions, students probably can more effectively learn to grasp the distinct properties of each disease even with a small sample of medical images from that class. Inspired by the learning process of humans, here we propose a simple yet effective method to embed attention into the learning process of deep neural network classifiers for intelligent diagnosis.

\subsection{Preliminary: Grad-CAM}
\textred{A two-step process must be performed to generate a gradient-based class discriminative activation map $F^c \in \mathbb{R}^{u\times v}$ for any given class $c$. First, the feature maps $A\in \mathbb{R}^{d\times u\times v}$ are computed immediately after the last convolutional layer, followed by the Rectified Linear Unit (ReLU) in the CNN. Secondly, the gradient score $\sigma{^{c}}$ of each class $c$, $y^c$, with respect to feature map activation of the last convolution layer $A^k$, i.e., $\frac{\partial y^c}{\partial A^k}$, needs to be calculated.}

\textred{Assuming the output of the final convolutional layer followed by ReLU is denoted by $A$, and the weight for the feature map of each channel (i.e., the last Linear Layer for ResNet50) is available, the final classification score $Y^c$ for a given class $c$ may be determined using the Equation~(\ref{eq:clsScore}).}

\begin{equation}\label{eq:clsScore}
	Y^c=\sum_k \underbrace{w_k^c}_{\text {class feature weights }} \overbrace{\frac{1}{Z} \sum_i \sum_j}^{\text {global average pooling }} \underbrace{A_{i j}^k}_{\text {feature map }}
\end{equation}
\textred{where $Z$ is the product of the height $u$ and width $v$ for each feature map $A_{k}$. For ResNet50~\cite{he2016deep}, used in our experiments, the feature map $A$ has $d=2048$ channels with a spatial size of $u=7$ and $v=7$. The weight $\sigma^c_{k}$ is calculated using the gradient computed from Equation~(\ref{eq:grad}).}

\begin{equation}\label{eq:grad}
	\sigma_k^c=\overbrace{\frac{1}{Z} \sum_i \sum_j}^{\text {global average pooling }} \underbrace{\frac{\partial Y^c}{\partial A_{i j}^k}}_{\text {gradients via backprop }}
\end{equation}
 
 \textred{The gradient can be computed through the backpropagation process, regardless of how many intermediate linear layers are present between the last convolutional layer and the classification layer. This holds for architectures such as VGG19~\cite{simonyan2014very}, which contains two intermediate linear layers. The gradient-based class discriminative feature map for a given class $c$, denoted as $F^c$, can be defined by Equation~(\ref{eq:featureMap}) with gradient and feature map.}

\begin{equation}\label{eq:featureMap}
	F^c=\sum_{k} \sigma_k^c A^k
\end{equation}

\textred{In this study, we only focus on the true positive class's feature map to simplify $F^c$ to $F$. We also normalize the feature map to the [0, 1] range and resize it back to [224, 224] for further attention loss computation. For clarity, we provide pseudocode for generating the feature map using Grad-CAM in Algorithm~\ref{alg:gradcam} in the Section~\ref{sec:appendix} Appendix.}
 
\subsection{CARE: class attention to regions of lesions}
We hypothesize that appropriate attention during learning would help neural network classifiers more effectively learn from small samples, particularly for rare diseases. Suppose the lesion regions of interest have been provided in advance for model learning, in the form of bounding boxes containing lesions. The human effort of providing bounding boxes is feasible for rare diseases because, quite often, only a small sample of images are available for each category, or alternatively, lesion regions could be automatically localized by saliency detection techniques (Section~\ref{sec:saliency}). Then, if there is a way to estimate the local regions on which the classifier focuses during image diagnosis, attention would be naturally embedded in those regions 
during classifier learning. Fortunately, such `visual focus' of a classifier on any input image can be conveniently estimated by a recently proposed visualization approach 
Grad-CAM~\cite{selvaraju2017grad}. Given a well trained classifier and an input image, Grad-CAM can provide a class specific feature activation map in which regions with higher activation contribute more to the classifier's output prediction being the specific class. Therefore, if the classifier attends to only the box bounded regions when diagnosing an image, the high activation regions from the Grad-CAM should also be within the bounded regions. In this sense, the spatial relationship (e.g., degree of overlap) between the high activation regions and the bounded image regions can be used to measure how well the classifier has attended to the bounded image regions.

Denote by $L_a$ the discrepancy between the high activation regions from Grad-CAM and the bounded image regions over all training data, then embedding attention during classifier learning can be realized by minimizing a new loss $L$ for the classifier,
\begin{equation}\label{eq:total}
L = (1 - \alpha) L_{c} +  \alpha L_{a}
\end{equation}
where  $L_c$ is the general cross entropy loss for the network classifier, and $L_a$, called \textit{attention loss}, which helps to drive the network to attend to box bounded image regions during training (Figure~\ref{fig:model}). $\alpha$ is a coefficient to balance the two loss terms. Considering the different influences of the inside box and outside box regions, the attention loss is further split into two items by
\begin{equation}\label{eq:La}
L_a = L_{in}+\lambda L_{out}
\end{equation}
where the inner loss $L_{in}$ helps the classifier increase the attention inside the bounding box, and the outer loss $L_{out}$ helps the classifier decrease the attention outside the bounding box. $\lambda$ is a coefficient to balance the two loss terms. In detail, for any training image with bounding box(es) provided, let $M_{in}$ denote a binary complement image in which all pixels inside the bounding box are set to $1$ and others to $0$, and in contrast, $M_{out}$ denote a binary mask image in which all pixels inside the bounding box are set to $0$, and any pixel outside the box is set to either $1$ or a positive value relevant to the distance between the pixel and the bounding box. \textred{In our implementation, we simplify setting the pixel outside the bounding box to $1$, since we assume that the bounding boxes completely enclose the lesions and an attention map outside the boxes should always incur a penalty}
Let $F$ denote the normalized feature activation map from Grad-CAM for the training image based on the current classifier. Then $L_{in}$ and $L_{out}$ (for one training image) can be defined as
\begin{equation}
\label{eq:Lin}
L_{in} = -\min(\frac{\Sigma_{i,j} M_{in}(i,j) \cdot F(i,j)}{\Sigma_{i,j}M_{in}(i,j)}, \tau)
\end{equation}
\begin{equation}\label{eq:Lout}
L_{out} = \frac{\Sigma_{i,j} M_{out}(i,j) \cdot F(i,j)}{\Sigma_{i,j}M_{out}(i,j)}
\end{equation}
Here, $M_{in}(i,j)$ represents the value at the position $(i,j)$ in the mask $M_{in}$, and similarly for $M_{out}(i,j)$ and $F(i,j)$. Equation~(\ref{eq:Lout}) represents the strength of feature activation outside the bounding box, while Equation~(\ref{eq:Lin}) would penalize the classifier if the highly activated area inside the bounding box is not large enough (i.e., when the percent of the weighted activated area $\frac{\Sigma_{i,j} M_{in}(i,j) \cdot F(i,j)}{\Sigma_{i,j}M_{in}(i,j)}$ is smaller than a predefined threshold $\tau$). Note that for notation simplicity, Equations~(\ref{eq:Lin}) and (\ref{eq:Lout}) are just for one single image. In fact, during training, the loss terms are calculated and averaged over all training images.

\input{figures/model}

One advantage of the proposed attention based approach is its independence of model structures. Therefore the CARE can be directly embedded in the training processing of any existing CNN classifiers, without alternating their model architectures. Also, the CARE framework is independent of existing approaches to handling data imbalance, therefore, can be directly combined to further improve classification performance.

\textred{We also provide a diagram for clarifying the training process. The presented diagram in Figure~\ref{fig:diagram} depicts the entirety of the process, encompassing data processing, model definition, and training methodology for the proposed model. The training of the proposed model, namely the CARE, can be categorized into two main stages: pretrain and attention based finetuning stages. The \textred{pretrain stage} is used to pretrain the model for generating a reasonable attention map by Grad-CAM, and the \textred{attention based finetuning stage} is trained with both the attention loss $L_a$ and the cross entropy loss $L_c$. During the pretrain stage, the CNN backbone and classifier are trained solely with data and corresponding labels. In the attention based finetuning stage, bounding boxes for the small class are generated via expert or automated techniques such as DSR and DeepLabV3, and subsequently processed to create training masks. The proposed model is then initialized with parameters from the pretrain stage and finetuned with the proposed attention loss alongside cross entropy loss. The initial feedforward and backward pass of the model produce an imprecise gradient-based class activation map. However, by utilizing the masks and a designed attention loss, attention outside the bounding boxes can be suppressed while directing attention toward the boxes, thus enabling the model to focus on the key aspects of the lesson. The full training process can be found in the pseudocode algorithm~\ref{alg:1} \textred{in Section~\ref{sec:appendix} Appendix. }}

\input{figures/diagram.tex}

\subsection{Automatic generation of bounding boxes} \label{sec:saliency}
While lesion regions can be annotated by human experts, it would be ideal if lesion regions could be automatically localized without help from human experts. Here we provide two possible solutions to the automatic localization of lesion regions, at least for some medical classification tasks. The first one is based on saliency detection, which can estimate the salient regions of images without any annotation information~\cite{li2013saliency}. In general, lesion regions often have a distinctive appearance (e.g., small darker regions in the image, as shown in Figure~\ref{fig:DSR_mask} compared to healthy backgrounds in images, and such distinctive appearance would be automatically found \textit{salient} by saliency detection techniques. Since the output of saliency detection is often a probability map representing the degree of saliency at each pixel location, binarization of the saliency map is applied based on a fixed or adaptive threshold, after which the compact bounding boxes surrounding the binarized salient regions can be easily obtained (Figure~\ref{fig:DSR_mask}). This study adopted a threshold value of 0.5, although the adaptive selection of the threshold could result in better performance.

\input{figures/DSRmask.tex}

Similar to the process of saliency detection for bounding box generation, the lesion segmentation model could be trained in advance to automatically estimate possible lesion regions from a healthy background in images, but under the strong assumption that annotations of lesion regions are available for some images from the same training set or another similar data set. \textred{There are many similar datasets available with segmentation labels. To generate bounding boxes for the small class, we first use an existing yet related dataset to train a DeepLabV3 segmentation model with Dice Loss, stopping when the model converges. Once trained, as shown in Figure~\ref{fig:deeplab_mask}, we freeze the model's parameters and use it to infer the data from the small class, producing a segmentation map where each pixel value ranges from 0-1. To obtain a binarized segmentation map, we further process the probability map with \textit{argmax}. Lesion regions are identified as positive values, while 0 is considered background. To prepare the segmentation map for CARE, we use \textbf{C}onnected-\textbf{C}omponent \textbf{L}abeling (CCL)~\cite{fiorio1996two,van2014scikit,wu2005optimizing} to remove small, unrelated lesions and obtain a rectangle containing the lesion. We set 1 for the lesion region and 0 outside the lesion region for the Mask $M_{in}$; the opposite process is followed for $M_{out}$. This process allows us to obtain the bounding boxes for the small class similar to human labeling.} Compared to unsupervised saliency detection, supervised lesion segmentation is much more limited in bounding box generation.

\input{figures/deeplabMask.tex}

%% file: figures/model.tex
\begin{figure}[htb]
    \centering
    \includegraphics[width=1\textwidth]{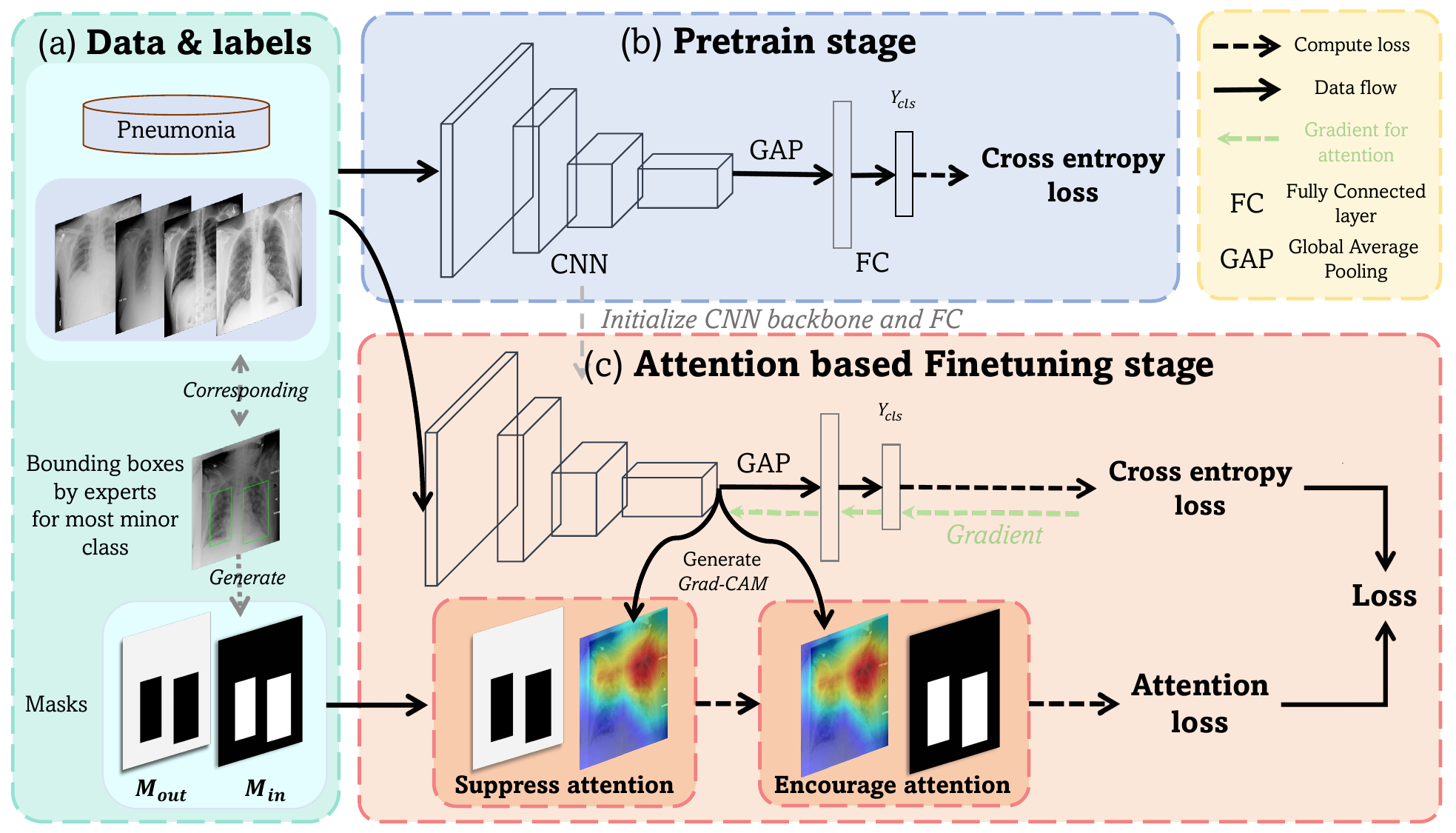}
    \caption{The framework of the proposed CARE method. Attention loss is designed to help the CNN model attend to lesion region for the minority category during model training. Bounding boxes representing lesions in images of only the minority category only need to be provided during training. The colorful heatmap is the activation map generated by Grad-CAM.}
    \label{fig:model}
\end{figure}

%% file: figures/diagram.tex
\begin{figure}[H]
    \includegraphics[width=1.0\linewidth]{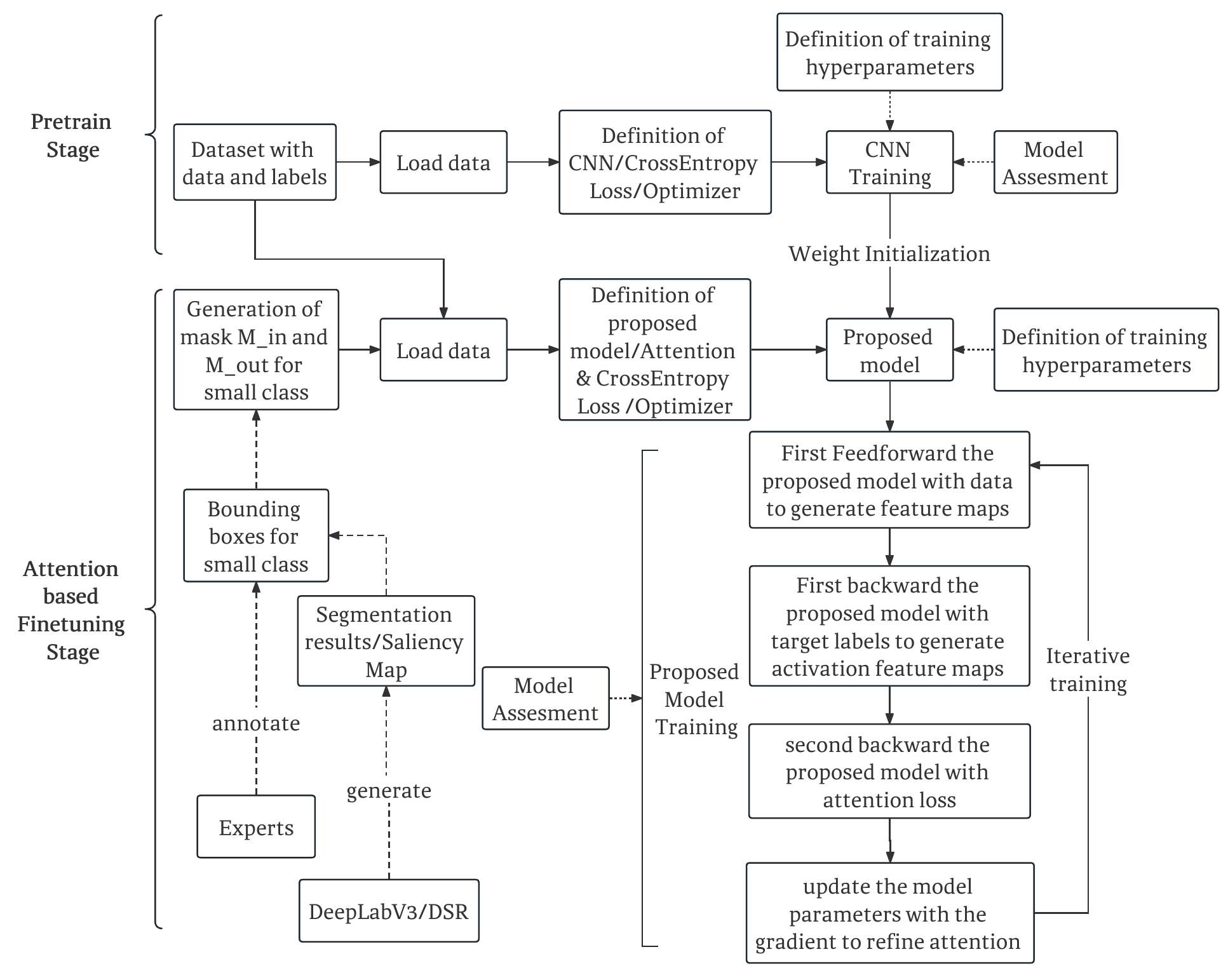}
    \caption{Diagram of the proposed method. The diagram can mainly be divided into two stages, including the pretrain stage and attention based finetune stage.}
    \label{fig:diagram}
\end{figure}

%% file: figures/DSRMask.tex
\begin{figure}[h]
	\centering
	\includegraphics[width=1.0\textwidth]{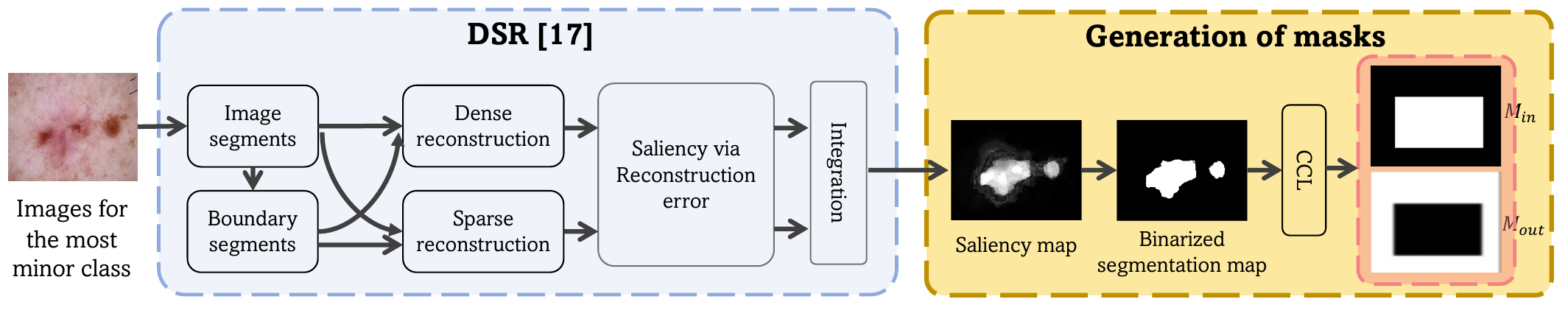}
   \caption{Masks $M_{in}$ and $M_{out}$ generation for the small class based on DSR with post processing. CCL represents Connected Component Labeling.}
	\label{fig:DSR_mask}
\end{figure}

%% file: figures/deeplabMask.tex
\begin{figure}[h]
	\centering
	\includegraphics[width=1.0\textwidth]{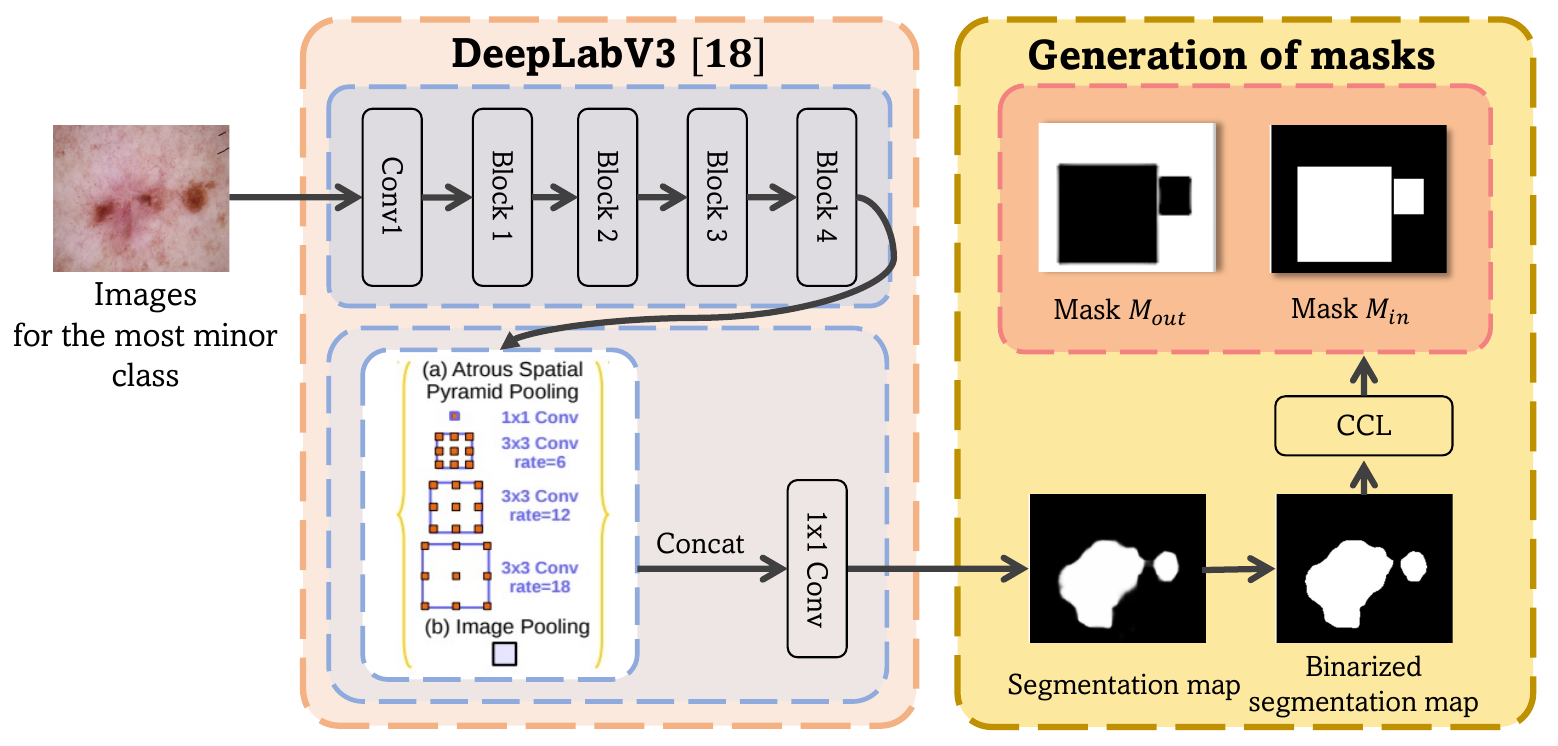}
   \caption{Masks $M_{in}$ and $M_{out}$ generation for the most minor class based on lesion segmentation model (DeepLabV3) with post processing. CCL represents Connected Component Labeling.}
	\label{fig:deeplab_mask}
\end{figure}

%% file: sec/experiments.tex
\section{Experiment}

\subsection{Experimental settings}
\subsubsection{Dataset}
Two medical image datasets were used to evaluate the proposed approach. One is the Skin Dataset provided by the ISIC2018 Challenge with seven disease categories~\cite{codella2018skin}, in which  6705 images are for Melanocytic nevus and only 115 images for Dermatofibroma, clearly having serious data imbalance among classes. One bounding box was generated for each image of the rare disease Dermatofibroma by the author and confirmed by a dermatologist. The other is the Pneumonia detection X-ray dataset with three categories \footnote{The original dataset comes from https://www.kaggle.com/c/rsna-pneumonia-detection-challenge, and part of the dataset was extracted for evaluation on data imbalance.}, including 8,851 `Normal' images, 105 `Lung Opacity' images, and 6,012 images of `No Lung Opacity/Not Normal'. Each `Lung Opacity' image was provided with one or multiple bounding boxes indicating the region of pneumonia. Although the original objective of this Chest X-ray data is for lesion detection, we used it for 3 class classifications, with the ground-truth bounding boxes used to evaluate the proposed approach. The number of `Lung Opacity' images is much smaller than the other two categories, being considered as the minority class in a data imbalance scenario. All images were resized to $224\times224$ pixels, with bounding boxes resized accordingly for the small sample class in each dataset. For each dataset, images are randomly split into a training set (80\%) and a test set (20\%) with stratification.

In the experiments, the training of CARE is divided into two stages. In the \textred{pretrain stage}, each backbone CNN classifier (i.e., the branch without the attention loss in Figure~\ref{fig:model}) used was pretrained firstly on ImageNet and then on the training set without the attention loss. The training at this stage is stopped when the cross entropy loss does not decrease anymore (normally within 200 epochs in our experiment). In the \textred{attention based finetuning} stage, attention loss was included to finetune the \textred{pretrain stage}'s classifier with the training set. $\alpha$ in Equation(\ref{eq:total}) was set to 0.5 unless otherwise mentioned. \textred{AdamW optimizer~\cite{loshchilov2017decoupled} was used in all experiments}, with an initial learning rate set at $0.0001$. $\lambda$ was empirically set to 0.5 for the X-ray dataset and 1.0 for the Skin dataset. 
More details of experimental settings were summarized in Table~\ref{tab:configs} for \textred{the pretrain and finetuning stages}.
In testing, each test image (without any bounding box) was fed to the CNN classifier for prediction. Note that the CARE approach needs no bounding box in testing.

\input{./tables/trainingConfig.tex}

To evaluate the performance of the proposed \textit{CARE}, two evaluation metrics suggested by the ISIC2018 Challenge were used here, including \textit{mean class accuracy} (MCA, i.e., average recall over all classes, treating contributions from each class equally)~\cite{chawla2002smote} and the area under the ROC curve (AUC)~\cite{hajian2013receiver,lobo2008auc}. Since AUC is general for binary classification, here for multi class classification tasks, AUC was first calculated for each class (treating all the other classes as a negative class) and then averaged. Besides, we use Recall for the smallest class (i.e., Dermatofibroma in Skin Dataset, and Lung Opacity in the Pneumonia dataset) to particularly highlight the performance of classifiers on the small sample class. 

 \textred{The architectures of the backbone used in our experiments like ResNet50~\cite{he2016deep} and VGG19~\cite{simonyan2014very} can be found in Figure~\ref{fig:resnet50_arch} and Figure~\ref{fig:vgg19_arch}, respectively.}

\input{figures/resnet50_arch.tex}
\input{figures/vgg19_arch.tex}

\subsection{Effectiveness of the proposed CARE}
\subsubsection{Baseline and comparison}
To test the effectiveness of the proposed approach, we compared CARE to three widely used strategies for handling data imbalance on both datasets, namely, 1) \textit{cost sensitive learning} denoted by CSL \cite{sun2007cost}, 2) \textit{focal loss} denoted by FL~\cite{lin2017focal}, a representative method of hard negative mining, and  3) \textit{data augmentation} (including rotation, flip and color jitter) denoted by DA~\cite{perez2017effectiveness}. We further tested our approach by embedding CARE into the three strategies, resulting in methods of CARE+CSL, CARE+FL, and CARE+DA. We also tested a baseline without using the visual attention loss in Figure~\ref{fig:model}. 
Table \ref{table:baseline} shows the comparison results on Pneumonia and Skin datasets with \textit{MCA} and \textit{AUC}, and on small classes with \textit{Recall}. It can be observed that CARE outperforms the baseline significantly in terms of \textit{Recall}, \textit{AUC}, and \textit{MCA}, in particular with a large margin on recall for the small sample class (31.1\% vs. 7.4\% on the Pneumonia dataset, and 73.9\% vs. 52.2\% on the Skin dataset). All three strategies (i.e., CSL, FL, and DA) perform better than the baseline without any treatment of data imbalance, which is expected. It is worth highlighting that adding CARE to each of CSL, FL or DA can boost the performances significantly w.r.t the use of each method alone; for instance, the Recall, AUC, and MCA of CARE+CSL on the Pneumonia dataset are respectively 45.0\%, 0.769 and 65.2\%, significantly better than CSL only. For the CSL method, additional experiments showed that varying loss coefficients for the minority class did not change the finding, i.e., CARE+CSL always performs better than CSL alone. This indicates that our approach is capable of boosting their performances significantly when plugged into the existing strategies for handling data imbalance. 
Figure~\ref{fig:skin_all_classes} demonstrates the detailed classification performance of each class. Compared with each baseline in \textred{Figure~\ref{fig:skin_all_classes} (a) and (b)}, our CARE method consistently improves the classification performance on minority classes (e.g., DF, VASC) while often slightly decreasing the performance on majority classes (e.g., NV, BKL). It has been widely observed in literature~\cite{cao2019learning} on imbalanced classification that the performance of one or a few majority classes would often become decreased a bit in accompanying the increased performance of minority classes.
\input{tables/baseline.tex}
\input{figures/skin_all_classes.tex}

\textred{In Figure~\ref{fig:process}, we illustrate the training process of the CARE(Ours) model on the Skin dataset (2nd row in Table 2) as an example. The table presents the loss curve, MCA curve, and recall of the most minor class on the training dataset, as well as the MCA and recall of the most minor class on the validation dataset. Figure~\ref{fig:process} (a) shows that the loss decreases as the training process advances. Figure~\ref{fig:process} (b) indicates that it may take up to 200 epochs for the model to converge and fit well on all data across all classes. Figure~\ref{fig:process} (c) suggests that training the model for more epochs can enhance the recall of the most minor class significantly, and MCA increases. However, due to the dataset's severe imbalance, the training curve may not be as stable as other balanced datasets, which is widely observed~\cite{cao2019learning}.}
\input{figures/process.tex}

In addition, two well known data imbalance strategies Siamese~\cite{hadsell2006dimensionality}~\footnote{\url{https://github.com/adambielski/siamese-triplet}} and Decouple~\cite{kang2019decoupling} were adopted for comparison on the Skin dataset. The decoupling method~\cite{kang2019decoupling} is a representative SOTA strategy dealing with data imbalance and has been re-implemented here. In detail, following the work~\cite{kang2019decoupling}, we freezed the parameters of the first 7 layers except for the last convolution block and classifier. As shown in Table~\ref{tab:siamese_decouple}, our proposed method boosts both methods, particularly on the small class DF and the overall classification performance over all classes.
\input{./tables/siamese_decouple.tex}

\subsubsection{Visual inspection}
To show the effect of the proposed attention loss, we visualize the classification activation maps of sample test images from both datasets \textred{in  Figure~\ref{fig:visual} (a) and (b).} For clarity, we also superimpose (ground truth) bounding boxes highlighting the lesion regions on the test images, provided along with the dataset (the Pneumonia dataset) or in-house annotation (the Skin dataset). Note that we did \textit{not} use any of those bounding boxes during testing, but here merely for visualization purposes. It can be observed that the activated regions (red regions in the second \textred{column of the Figure~\ref{fig:visual} (a) and the second column of Figure~\ref{fig:visual} (b)}) without using the proposed attention loss (Eq. \ref{eq:La}) were deviated from or not focused on lesion regions, while those (\textred{third column of Figure~\ref{fig:visual} (a) and last columns of Figure~\ref{fig:visual} (b)}) produced by CARE localized the lesion regions well.
These results on the test images reveal that the CARE model could have learned to focus on lesion regions when analyzing new images, through attention loss optimized during training. 
\textred{On the other hand, as shown in the last two rows of Figure~\ref{fig:visual}(b), 
\textit{CARE} did not always help the model focus on the annotated lesion region. However, compared with the results \textit{without CARE} (second last column), \textit{CARE} help the model focus on more reasonable areas like the chest and heart where the shape and texture of the area look very similar to the actual lesion. This again indicates that \textit{CARE} tries to help the model capture the actual lesion to make the classification decision. We found that failure cases more likely appear in the Pneumonia dataset than the Skin dataset, probably because the actual lesions in the Pneumonia dataset are relatively small and often similar to other image regions. Replacing Grad-CAM with a more precise class activation map generation method during model training could improve the model's performance.   
}

\input{figures/comparisionSkinPneu.tex}

\subsection{Robustness of CARE}
\subsubsection{Flexibility with model architecture}
Our proposed CARE framework is independent of model structures. To show this, we tested variants of our CARE framework built with two different widely used CNN architectures: ResNet~\cite{he2016deep} and VGG19~\cite{simonyan2014very}. For ResNet, we further tested the different number of layers (18, 50, and 152 layers), from shallow to very deep. VGG19 uses a 19 layered structure. Thus in total, we have four backbones of CNN architectures: ResNet18, ResNet50, ResNet152, and VGG19. With each backbone, we compared the performance of the resulting CARE model (denoted by X(CARE), with X representing the name of the backbone) and the original backbone network. From Table \ref{tab1e:flex}, it can be observed that different \textit{original} backbone architectures perform differently, among which VGG19 performs the best. For each of the backbone, its CARE version consistently outperforms the original network in terms of both recall and MCA. This confirms the flexibility of the proposed CARE in the combination of various model architectures.
\input{tables/flex.tex}

\subsubsection{Tolerance to bounding box accuracy}
It is noted that the training of our model needs bounding box annotations for the smallest class. For many rare or uncommon diseases (such as Dermatofibroma in this study), the \textred{bounding box} annotation effort for the lesion regions in the minority class is usually very small compared to that of accurate boundary pixel level annotations. Even though, there might exist inter- or intra-observer variations in annotation. The bounding boxes used thus far are tightly around the lesion region. To relax this requirement, we vary the bounding boxes by scaling at 0.7, 0.9, 1.0, and 1.1, and test the robustness of our approach to such scaling. \textred{In detail, we utilized manually labeled bounding boxes as the reference with a scale of 1.0 and varied the size of the bounding boxes that covered  the lesion region using different ratios before feeding the model. Specifically, we increased and decreased the size of the original bounding box that encompasses the lesion area by factors of 0.7 to 1.1 to either encompass a smaller or larger region of the lesion in the images.}. Figure~\ref{fig:bbox} shows the performance of our model at different scaling. It can be seen that the performance remains stable when the box size changes within a reasonable range, for instance, from 0.7 to 1.1. This indicates that our approach is largely tolerant to the sizes of bounding boxes, i.e., the bounding boxes do not need to be tight around the lesions, with the flexibility of using a looser bounding box, which would require less annotation effort from human experts.
\input{figures/bbox.tex}

\subsubsection{Effect of $\alpha$}
We further conducted a set of experiments to evaluate the effect of the coefficient parameter $\alpha$ of the attention loss term using ResNet50(CARE), and the results are shown in Table~\ref{table:alpha}. It can be observed that the performance of the model remains consistently better than the baseline (with $\alpha=0$) when $\alpha$ changes within a reasonably large range, for instance, from 0.3 to 0.9, and particularly the MCA performance remains stable with varying $\alpha$ values on both datasets. This indicates that our model is insensitive to the choices of the value of $\alpha$ in improving the classification performance. 
\input{tables/alpha.tex}

\subsubsection{Effect of $\tau$}
In addition, the effect of the threshold hyperparameter $\tau$ in the loss term $L_{in}$ was evaluated by varying its values from 0.2 to 1.0 with the ResNet50(CARE). Table~\ref{table:tau} shows that, although the classification performance vary with different threshold values, they all outperform the baseline model (without using CARE). Figure~\ref{fig:different_tau} shows with increasing value of the threshold, the model would be trained to attend to larger lesion regions during diagnosis, which is as expected and confirms the role of the loss term $L_{in}$. 
$\tau$ is a threshold to discourage the attention to fulfill the whole bounding box. Since lesions of some diseases such as Skin tumors may occupy most of the bounding box, and some other diseases may only occupy a relatively smaller area within the bounding box indicating a smaller $\tau$ should be used, the choice of $\tau$ may be different on different datasets. This sensitivity study shows that model performance remains stably high within a large range of $\tau$ values on both datasets. In practice, $\tau$ was empirically set based on a small internal validation set from each dataset, and many other choices would result in similar performance based on the sensitivity study.

\input{tables/tau.tex}
\input{figures/differentTau.tex}

\subsubsection{Effect of $\lambda$}
For the coefficient $\lambda$,  Table~\ref{tab:lambda} shows that, when $\lambda$ varies from 0.7 to 1.3 (2nd to 4th rows), the model achieves similar high performance in MCA over all classes and improved performance in recall for the smallest class DF, again confirming the robustness of the proposed method.

\input{./tables/lambda.tex}

\subsection{CARE with automated bounding box generation}
This experiment tests whether automatically estimated bounding boxes would also help improve the classification performance on the small sample class of skin images. While there exist multiple solutions to automatic bounding box generation, here as two examples, we adopted a traditional unsupervised saliency detection method DSR~\cite{zhang2015exploiting} and a well known supervised segmentation model DeepLabV3~\cite{chen2017rethinking}. 
\textred{To use the segmentation model in our CARE method, we need to train a segmentation model using certain dermoscopic image dataset, with lesion region annotations. Here the Skin dataset from the ISIC2018 Skin Segmentation Challenge, which includes 2,594 dermoscopic images with lesion region annotations, was used to train the DeepLabV3 model. It is important to note that this dataset is \textit{different from the skin disease types used for the classification task} but is still appropriate for our CARE method since these images share similar edge and texture features and backgrounds. }
Note that all the small class images used for classification were not included in the segmentation training set. Figure~\ref{fig:autobbox} shows examples of bounding boxes automatically generated by the two methods.

\input{figures/autobox.tex}

With the automatically estimated bounding boxes, the two models trained by the CARE framework, i.e., CARE(DSR) and CARE(DeepLabV3) with the backbone ResNet50 in Table~\ref{table:autobbox}, outperform the baseline model without using CARE, particularly on the smallest class. This confirms the effect of the CARE framework on performance boosting even with automatically estimated bounding boxes. It can also be observed from Table~\ref{table:autobbox} that the network model CARE(GT) trained with the manually labeled bounding boxes performs the best. Figure~\ref{fig:autocare_visual} shows the corresponding visualization results.
\input{tables/autobox.tex}
\input{figures/autocareVisual.tex}
This is reasonable because the manually labeled bounding boxes are often more accurate and less erroneous compared to those by the saliency detection or segmentation methods. As part of future studies, better methods of estimating lesion regions could be developed and applied here. 
The visual illustration in Figure~\ref{fig:autocare_visual} also shows that the network models trained with automatically estimated bounding boxes (\textred{Figure~\ref{fig:autocare_visual} (d) and (e)}) can also well attend to lesion regions during diagnosis, although sometimes the attention is slightly worse (\textred{Figure~\ref{fig:autocare_visual} (d)}) than that from the network model trained with the manually labeled bounding boxes (\textred{Figure~\ref{fig:autocare_visual} (a)}).

Similar to the above tests of combining the CARE framework with existing approaches to data imbalance issue (Table~\ref{table:baseline}),  Table~\ref{table:autobboxcombine} clearly shows that using the automatic generated bounding boxes, the combination of CARE with CSL or FL also significantly improved the performance of the network model trained with CSL or FL alone. In particular, the improvement on the smallest class (Recall) is more significant. This further supports that even with the possibly inaccurate estimation of bounding boxes, the CARE framework is effective to further improve classification performance when combined with existing data imbalance strategies.   
\input{tables/autoboxCombine.tex}

\subsection{Increasing classes of masks in CARE}
The algorithm is not limited to the class with the smallest sample size. The three smallest categories of the Skin dataset are DF, VASC, and AKIEC, respectively, having 115, 142, and 327 images. Segmentation model DeepLabV3 was used to generate masks for the three classes of training data, and the generated masks were double checked by medical experts to ensure most masks were reasonably good enough. As shown in Table~\ref{tab:more_minorities}, applying the proposed method to DF, DF+VASC, and DF+VASC+AKIEC can clearly improve the recall of the small classes (56.52$\rightarrow$89.96, 89.65$\rightarrow$96.55, 68.18$\rightarrow$74.24) and the MCA over all classes compared to the baseline (cost sensitive learning, CSL). 
It can also be observed that the performance improvement on the smallest class DF becomes smaller when masks are provided for the other small (but relatively larger) classes VASC and AKIEC, during model training. This is probably because attention is biased toward the relatively larger classes VASC and AKIEC during model training by our CARE method.
    
\input{./tables/increase.tex}

%% file: tables/trainingConfig.tex
\begin{table*}[h]
    \centering
    \caption{Details of experimental setting on the Skin dataset. `-' denotes that the hyperparameter is not used in the \textred{pretrain stage}.}
    \scriptsize
    \label{tab:configs}
    \begin{tabular}{c|cccccccc}
        \toprule
         Config & Learning Rate & Batch Size & Epochs & Activation Feature Map & $\alpha$ & $\lambda$ & $\tau$\\
         \midrule
         Pretrain Stage & 1e-4 & 96 & 200 & - & - & - & - \\
         Finetune Stage & 1e-4 & 96 & 200 & $224\times 224$ & 0.5 & 1 & 0.7 \\
         \bottomrule
    \end{tabular}
\end{table*}

%% file: figures/resnet50_arch.tex
\begin{figure*}[ht]
	\centering
	\includegraphics[width=1.0\linewidth]{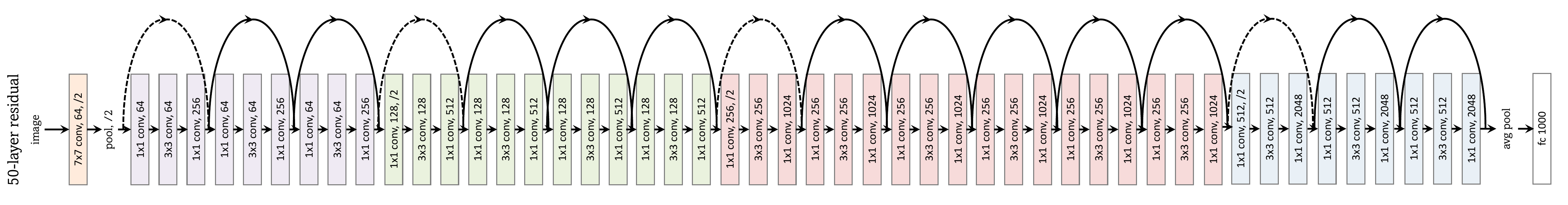}
	\caption{Architecture of the ResNet50~\cite{he2016deep} backbone.}
	\label{fig:resnet50_arch}
\end{figure*}

%% file: figures/vgg19_arch.tex
\begin{figure*}[ht]
	\centering
	\includegraphics[width=1.0\linewidth]{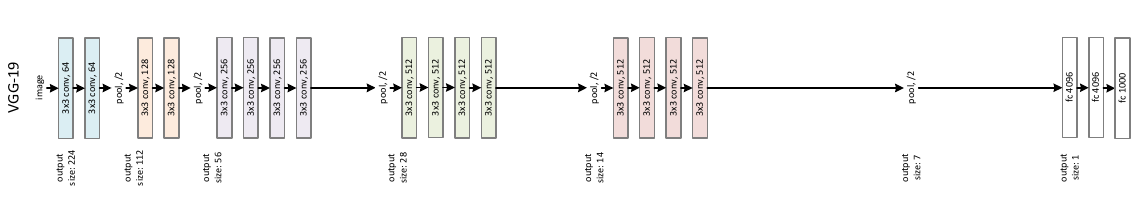}
	\caption{Architecture of the VGG19~\cite{simonyan2014very} backbone.}
	\label{fig:vgg19_arch}
\end{figure*}

%% file: tables/baseline.tex
\begin{table}[tb]
    \centering
    \caption{Comparison between baselines and our method on the two commonly used imbalanced medical image datasets with backbone ResNet50. }
    \scriptsize
    \begin{tabularx}{0.95\textwidth}{*{7}{ccccccc}}
    	\toprule
      	\multirow{2}{*}{Method} & \multicolumn{3}{c}{Pneumonia dataset} & \multicolumn{3}{c}{Skin dataset} \\
      	\cmidrule(l){2-4} \cmidrule(l){5-7}
        & \tabincell{c}{Recall of the\\most minor class} & AUC & MCA & \tabincell{c}{Recall of the\\most minor class} & AUC & MCA \\ 
        \midrule
      	baseline & 7.4 & 0.663 & 56.8 & 52.2 & 0.962 & 74.1\\
      	CARE (ours) & \textbf{31.1}  & \textbf{0.741} & \textbf{63.3} & \textbf{73.9} & \textbf{0.973} & \textbf{78.2}\\	
		\hline
      	CSL & 11.1 & 0.704 & 57.9 & 56.5 & 0.981 & 77.2\\
      	CARE+CSL(ours) & \textbf{45.0}  & \textbf{0.769} & \textbf{65.2} &  \textbf{65.2} & \textbf{0.986} & \textbf{81.0}  \\
      	\hline
      	FL & 11.1 & 0.758 & 58.4 & 52.2 & 0.918 & 71.5\\
      	CARE+FL (ours)  & \textbf{49.4} & \textbf{0.769} & \textbf{66.7}  & \textbf{60.9} & \textbf{0.933} & \textbf{73.9} \\	
      \hline
       DA & 20.1 & \textbf{0.805} & 59.6 & 56.6 & 0.862 & 54.4 \\
       CARE+DA(ours) &\textbf{45.2} & 0.796 &\textbf{66.0} & \textbf{60.3} & \textbf{0.885} & \textbf{56.2} \\
       \bottomrule
    \end{tabularx}
    \label{table:baseline}
\end{table}

%% file: figures/skin_all_classes.tex
\begin{figure}[H]
	\centering
	\includegraphics[width=1.0\linewidth]{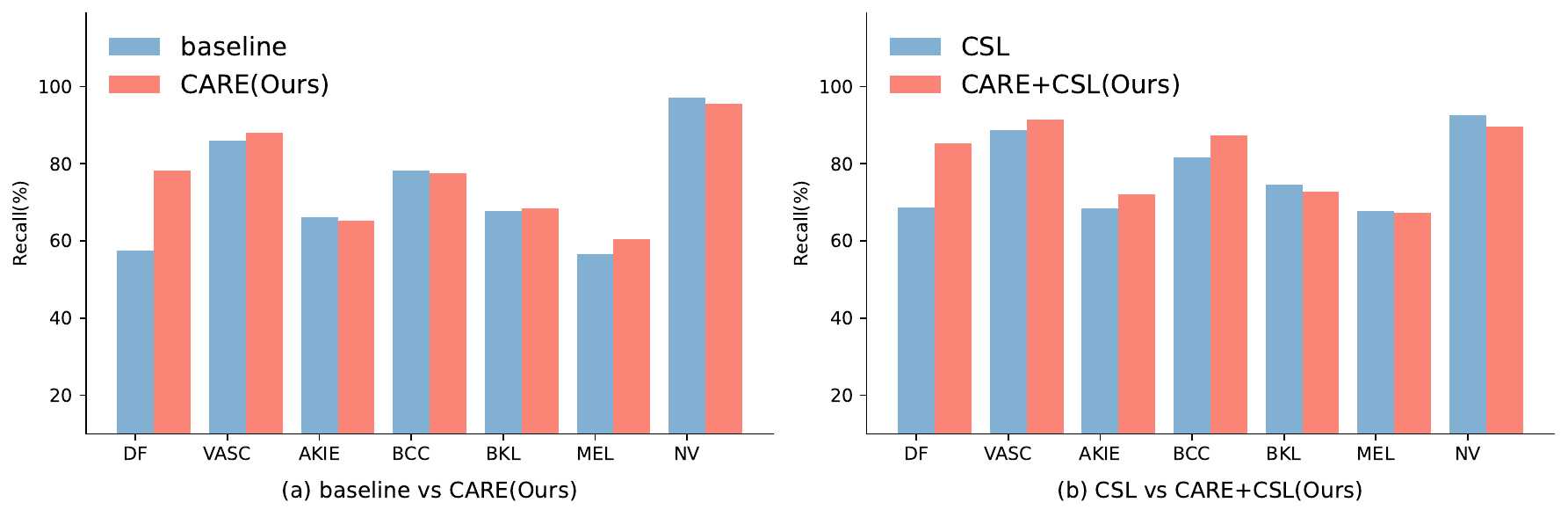}
	\caption{Comparison between baselines and our method in each class on the Skin dataset.}
	\label{fig:skin_all_classes}
\end{figure}

%% file: figures/process.tex
\begin{figure}[H]
    \centering
    \includegraphics[width=1\textwidth]{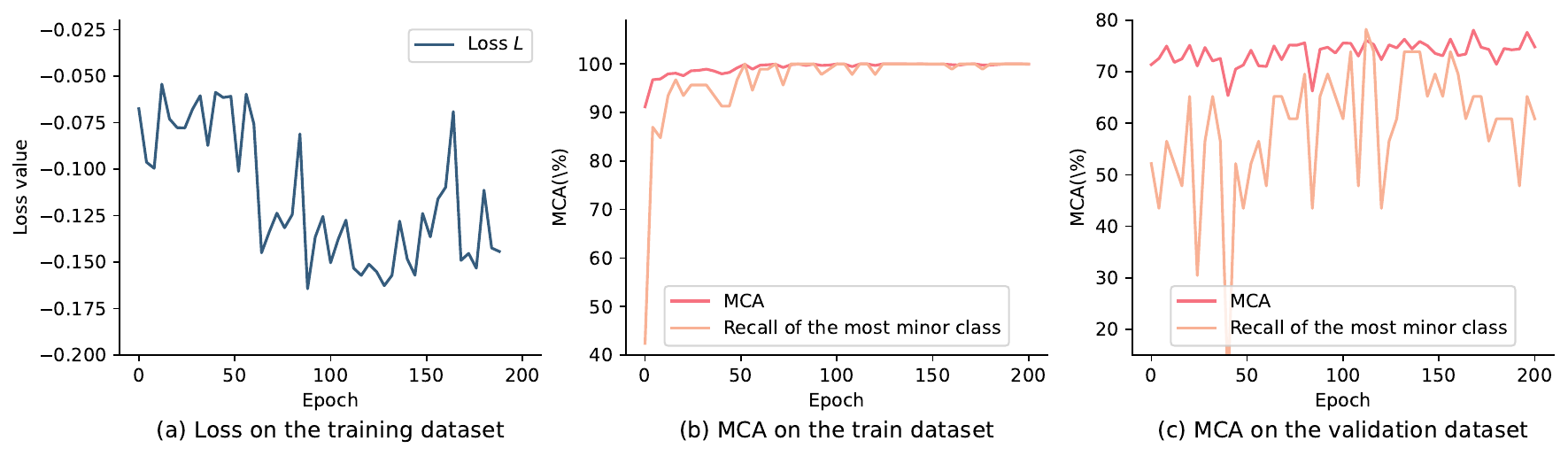}
    \caption{The training process details for the CARE(Ours) in Table 2. (a) displays the changes in the total loss throughout the training process. (b) shows the changes in the MCA and recall of the most minor class on the training dataset during the training process. (c) presents the changes in the MCA and recall of the most minor class on the validation dataset. }
    \label{fig:process}
\end{figure}

%% file: tables/siamese_decouple.tex
\begin{table*}[h]
		\centering
		\caption{Comparison with the Siamese and the Decoupling methods on the Skin dataset.} 
		\label{tab:siamese_decouple}
		\footnotesize
		\begin{tabular}{ccccccccc}
			\toprule
			\multirow{2}{*}{Method} & \multirow{2}{*}{\textbf{MCA}} & \multicolumn{7}{c}{ \textbf{Recall} } \\
		\cline { 3 - 9 }  & & DF & VASC & AKIEC& BCC & PBK & MEL & NV \\		
			\midrule
			Siamese & 56.9 & 21.7 & 51.7 & 53.0 & 67.9 & 60.0 & 45.7 & 98.3\\
			Siamese+CARE(ours) & {68.9} & {65.2} & 68.9 & 48.4 & 70.8 & 65.4 & 70.8 & 93.8\\
            \hline
			Decouple & 73.9 & 65.2 & 82.7 & 57.5 & 81.5 & 76.3 & 63.2 & 91.9\\
			Decouple+CARE (ours) & {78.3} & {73.9} & 89.6 & 62.1 & 83.4 & 74.0 & 73.0 & 92.2\\
			\bottomrule
		\end{tabular}
\end{table*} 

%% file: figures/comparisionSkinPneu.tex
\begin{figure}[H]
	\centering
	\includegraphics[width=1\textwidth]{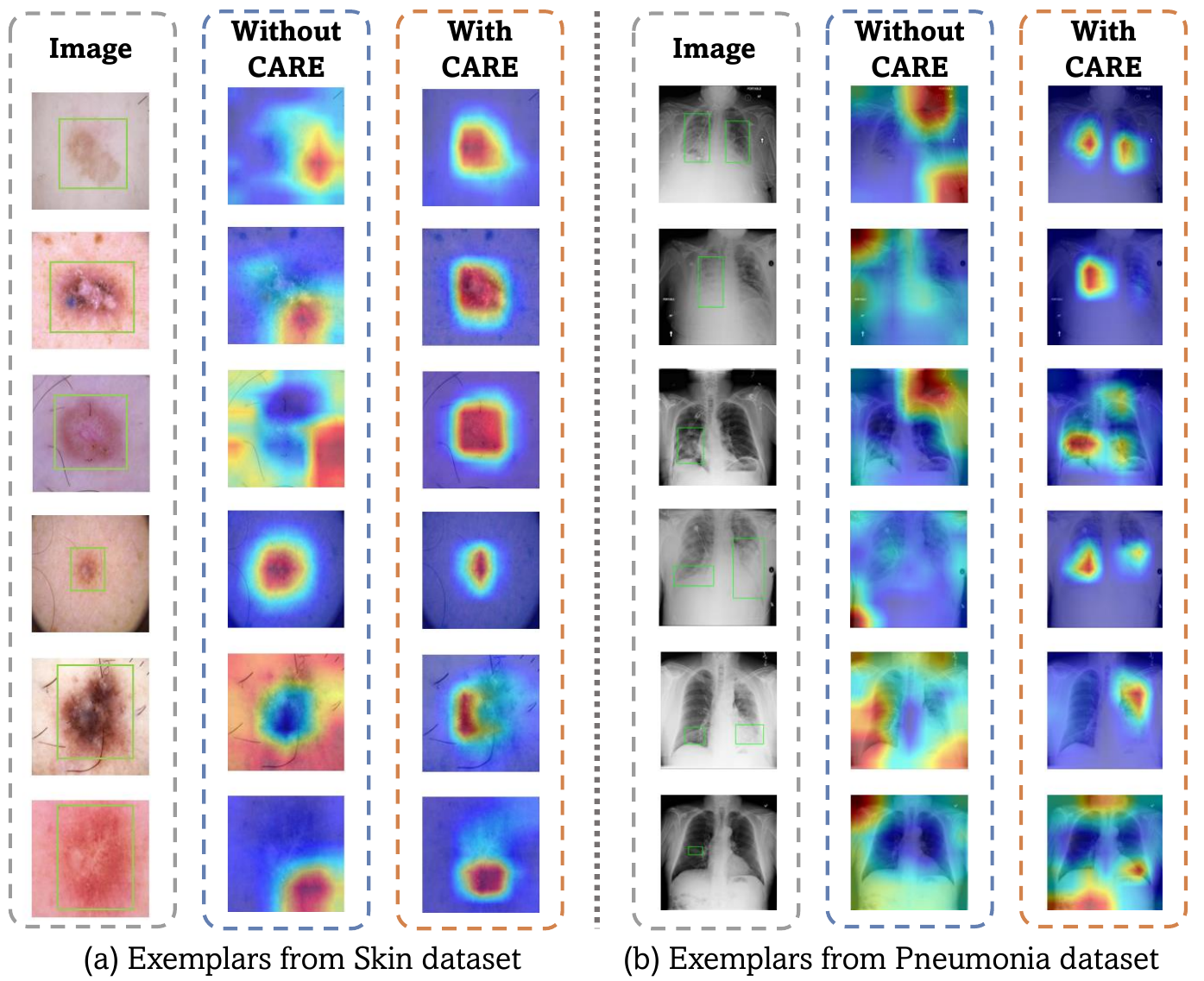}
	\caption{Visualization of activation maps with and without using CARE on (a) the Skin Dataset and (b) the Pneumonia Dataset. First column: test images superimposed with bounding boxes; the Second and third columns: activation maps without and with CARE attention loss, respectively. Similarly for the last three columns. 
	{We use Grad-CAM to estimate class activation maps (best viewed in color)}}
	\label{fig:visual}
\end{figure}

%% file: tables/flex.tex
\begin{table}[tb]
    \centering
    \caption{
    {Performance of CARE with various model architectures on two commonly used imbalanced medical image datasets. X(CARE) means that the CARE has the backbone X, e.g., ResNet18(CARE) represents the CARE with the backbone model ResNet18. It's worth noting that all models apply CSL in this table.
    }
    } 
		\footnotesize
        \scriptsize
       \begin{tabularx}{0.95\textwidth}{*{7}{ccccccc}}
    	\toprule
      	\multirow{2}{*}{Backbone} & \multicolumn{3}{c}{Pneumonia dataset} & \multicolumn{3}{c}{Skin dataset} \\
      	\cmidrule(l){2-4} \cmidrule(l){5-7}
        & \tabincell{c}{Recall of the\\most minor class} & AUC & MCA & \tabincell{c}{Recall of the\\most minor class} & AUC & MCA \\ 
        \midrule
      	ResNet18 & 15.2 & 0.581 & 57.8 & 60.9 & \textbf{0.990} & 74.1\\
      	ResNet18(CARE) & \textbf{25.5} & \textbf{0.710} & \textbf{58.8} & \textbf{73.9} & 0.969 & \textbf{76.2}\\	
		\hline
      	ResNet50 & 11.1 & 0.704 & 57.9 & 56.5 & 0.953 & 77.7\\
      	ResNet50(CARE) & \textbf{45.0}  & \textbf{0.769} & \textbf{65.2} &  \textbf{65.2} & \textbf{0.986} & \textbf{81.0}  \\
      	\hline
      	ResNet152 & 11.4 & 0.747 & 59.1 & 61.9 & 0.914 & 80.5\\
      	ResNet152(CARE) & \textbf{31.3} & \textbf{0.749} & \textbf{63.8}  & \textbf{72.2} & \textbf{0.966} & \textbf{81.9} \\	
      \hline
       VGG19 & 25.8 & \textbf{0.873} & 61.7 & 47.8 & 0.881 & 70.2 \\
       VGG19(CARE) &\textbf{41.2} & 0.871 &\textbf{64.3} & \textbf{56.5} & \textbf{0.933} & \textbf{72.8} \\
       \bottomrule
    \end{tabularx}
    \label{tab1e:flex}
\end{table}

%% file: figures/bbox.tex
\begin{figure}[H]
	\centering
	\includegraphics[width=0.6\textwidth]{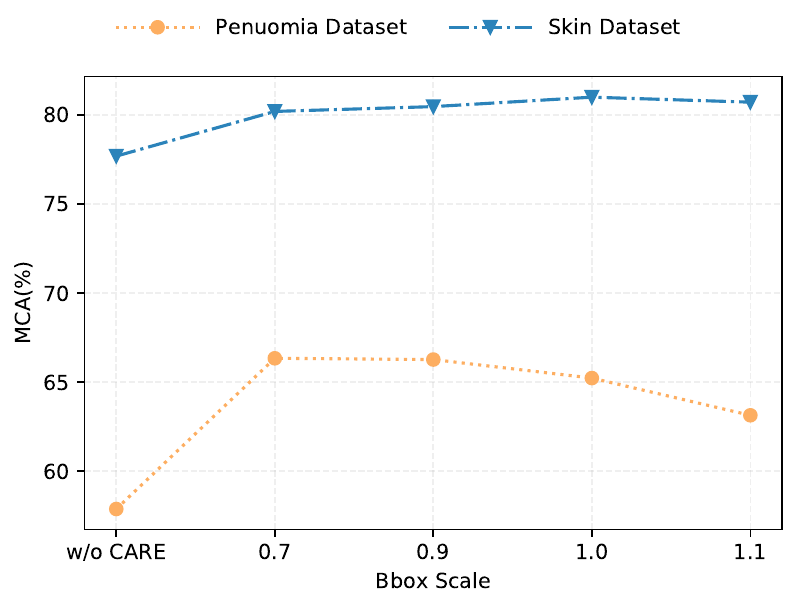}
  \caption{Robustness to scales of bounding box, tested with ResNet50(CARE) on the Skin dataset.}
	\label{fig:bbox}
\end{figure}

%% file: tables/alpha.tex
\begin{table}[th]
    \centering
    \caption{Effect of the coefficient $\alpha$ on classification performance with ResNet50(CARE). Note that all models here apply CSL.} 
     \footnotesize
    \begin{tabular}{*{5}{ccccc}}
            \toprule
      \multirow{2}{*}{$\alpha$} & \multicolumn{2}{c}{Pneumonia dataset} & \multicolumn{2}{c}{Skin dataset} \\
      \cmidrule(l){2-3} \cmidrule(l){4-5}
        & \tabincell{c}{Recall of the\\most minor class} & MCA & \tabincell{c}{Recall of the\\most minor class} & MCA \\ 
            \midrule
            0 & 11.1 & 57.9  & 56.5 & 77.7\\
            0.1  & 28.8   & 63.8  & 55.2 & 78.8 \\ 
            0.3  & 38.3  & 64.4  & {65.2} & 79.9  \\ 
            0.5  & 45.0 & {65.2} & 65.2  & {80.3} \\
            0.7  & 42.6   & 64.3 & 65.2 & 80.2\\ 
            0.9  &{50.7}   & 65.2  & 65.1 & 80.2\\ 
      \bottomrule
    \end{tabular}
        \label{table:alpha}
\end{table}

%% file: tables/tau.tex
\begin{table}[tb]
   \centering
   \caption{Effect of the hyper-parameter $\tau$ on classification performance.}
        \footnotesize
    \begin{tabular}{ccccc}
    \toprule
    \multicolumn{1}{c}{\multirow{2}[4]{*}{$\tau$}} & \multicolumn{2}{c}{Penuomia Dataset} & \multicolumn{2}{c}{Skin Dataset} \\
  \cmidrule(l){2-3} \cmidrule(l){4-5}          & \tabincell{c}{Recall of the\\most minor class} & MCA & \tabincell{c}{Recall of the\\most minor class} & MCA \\
    \midrule
    \multicolumn{1}{c}{baseline} & 11.1 & 57.9 & 56.5 & 77.7 \\
    0.2 & 33.3 & 64.9 & 73.9 & 79.1 \\
    0.4   & {51.8} & {68.9} & {69.6} & 78.8 \\
    0.5   & 45.0 & 65.2 & 65.2 & {81.0} \\
    0.7   & {51.9} & 67.4  & 65.2 & 80.2 \\
    1     & 37.0 & 64.0 & 78.3 & 79.6 \\
    \bottomrule
    \end{tabular}%
 \label{table:tau}%
\end{table}%

%% file: figures/differentTau.tex
\begin{figure}[th]
	\centering
	\includegraphics[width=1\textwidth]{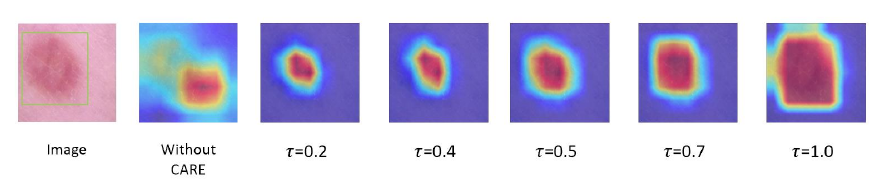}
	\caption{Exemplar effect of the hyper-parameter $\tau$. Higher $\tau$ values cause larger attended regions on the test image.}

\label{fig:different_tau}
\end{figure}

%% file: tables/lambda.tex
\begin{table*}[h]
		\centering
		\caption{Effect of the coefficient $\lambda$ on classification performance on the Skin dataset. The 1st row represents the results of CSL without CARE.}
        \footnotesize
		\label{tab:lambda}
		\begin{tabular}{c|cccccccc}
		\toprule
		\multirow{2}{*}{$\lambda$} & \multirow{2}{*}{{\textbf{MCA}}} & \multicolumn{7}{c}{ \textbf{Recall} } \\
		\cline { 3 - 9 }  & & DF & VASC & AKIE & BCC & PBK & MEL & NV \\		
		\hline 
		\rowcolor{mygray}- & 75.9 & 56.5 & 93.1 & 66.6 & 80.5 & 76.8 & 64.1 & 94.0 \\
		0.7 & 79.95 & 60.8 & 96.5 & 78.7 & 87.3 & 76.8 & 67.2 & 92.0 \\
		1.0 & {80.6} & {86.9} & 89.6 & 72.7 & 86.4 & 68.1 & 71.7 & 88.2 \\
		1.3 & 79.94 & 69.5 & 96.5 & 68.1 & 88.3 &  70.4 & 78.4 & 87.9\\
	\bottomrule
\end{tabular}
	\end{table*}

%% file: figures/autobox.tex
\begin{figure}[h]
\centering
	\includegraphics[width=1.0\textwidth]{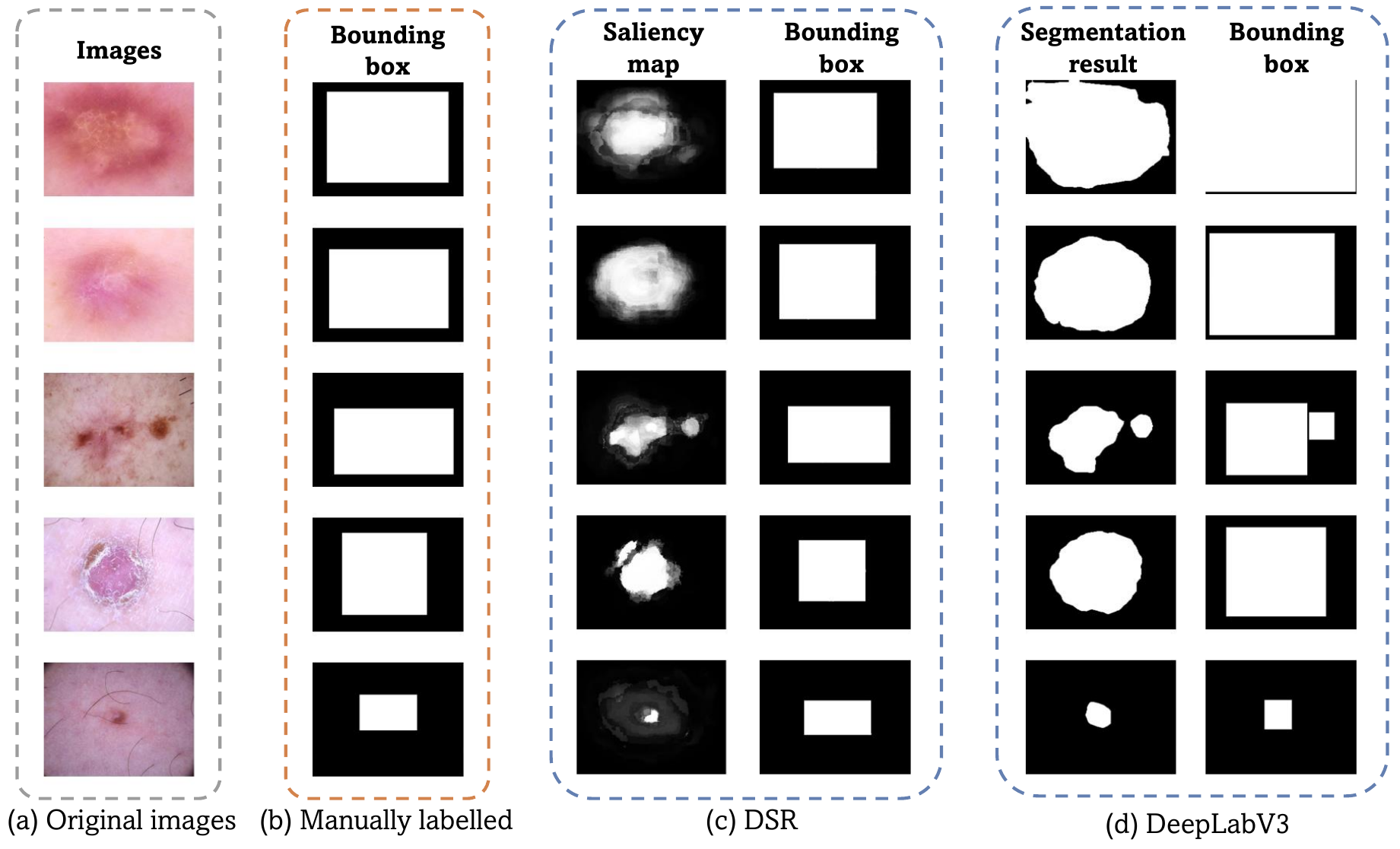}
  \caption{Automated generation of bounding boxes for the minority class in the Skin Dataset. From left to right: (a) original image, (b) manually labeled bounding box, (c) saliency map by DSR, bounding box based on DSR, (d) segmentation result by DeepLabV3, bounding box based on segmentation.}
	\label{fig:autobbox}
\end{figure}

%% file: tables/autobox.tex
\begin{table}[ht]
  \centering
  \caption{Performance of CARE models on the Skin Dataset, with bounding boxes from different ways. CARE(X) denotes the CARE model with bounding boxes from X.}
        \footnotesize
    \begin{tabularx}{0.95\textwidth}{ccccc}
    \toprule
    Method & Baseline & CARE(GT) & CARE(DSR) & CARE(DeepLabV3)\\
    \midrule
    \textbf{MCA} & 74.13 & \textbf{78.17} & 77.02 & 75.61 \\
    \tabincell{c}{Recall of the\\most minor class} & 52.17 & \textbf{73.91} &  65.22 &  60.87 \\
    \bottomrule
    \end{tabularx}%
  \label{table:autobbox}%
\end{table}%

%% file: figures/autocareVisual.tex
\begin{figure}[H]
	\centering
	\includegraphics[width=1.0\textwidth]{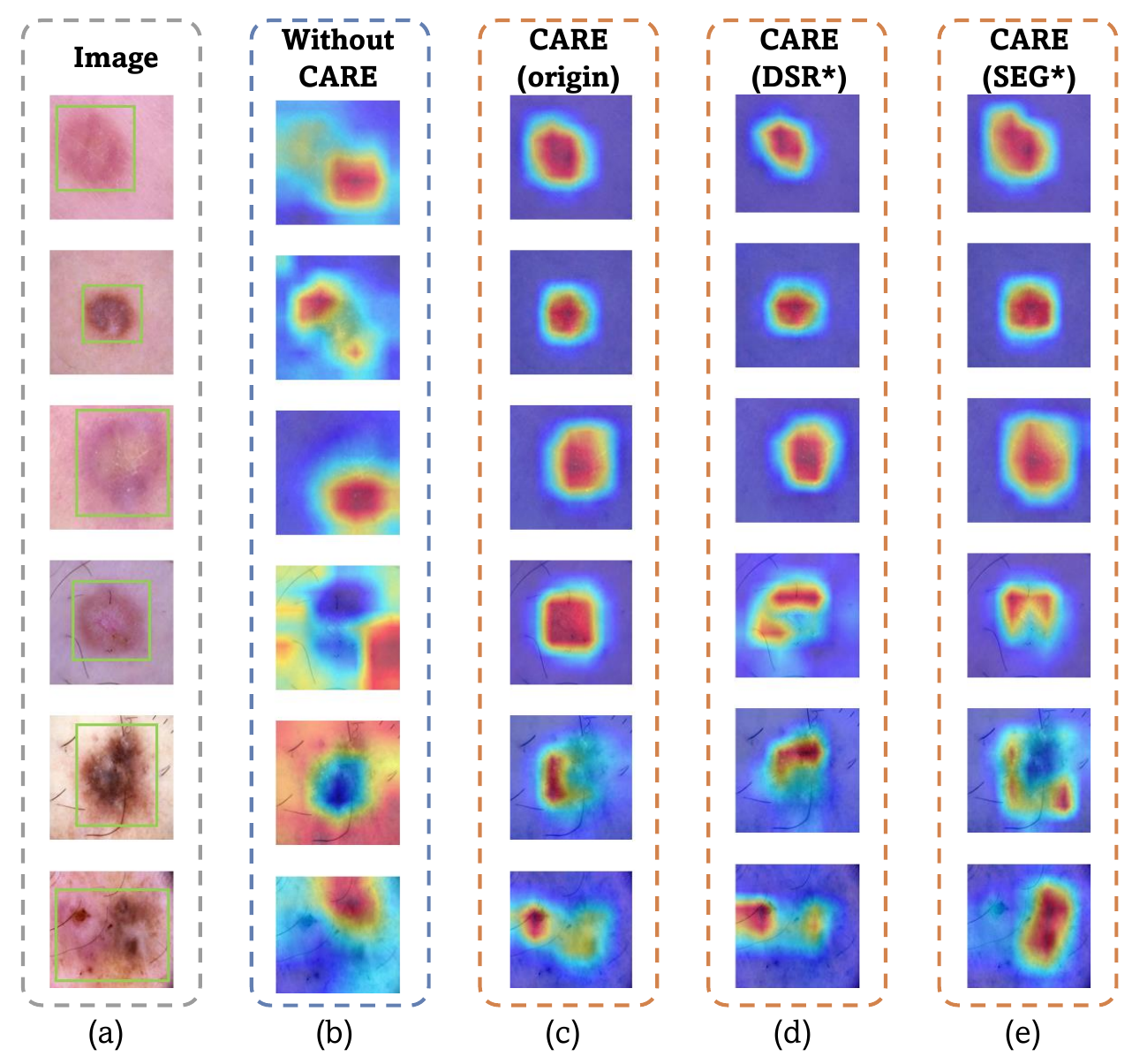}
  \caption{Visualization of activation maps with different versions of CARE. From first to last column: (a) original images superimposed with bounding boxes, (b) activation maps without using CARE attention loss, (c) activation maps with CARE(GT), (d) CARE(DSR), and (e) CARE(DeepLabV3), respectively. We use Grad-CAM to generate the class activation maps (best viewed in color).
  }
	\label{fig:autocare_visual}
\end{figure}

%% file: tables/autoboxCombine.tex
\begin{table}[ht]
	\centering
	\caption{Performance of the ResNet50 models on the skin dataset when trained in the CARE framework together with CSL (Cost Sensitive Learning) and FL (Focal Loss) respectively. Bounding boxes used in CARE were estimated by the saliency detection method DSR or the segmentation model DeepLabV3.} 
    \tiny
	\begin{tabular}{ccccccc}
	\toprule
    \tabincell{c}{Method} & \tabincell{c}{CSL} & \tabincell{c}{CSL+\\CARE(DSR)} & \tabincell{c}{CSL+\\CARE(DeepLabV3)} & \tabincell{c}{FL} & \tabincell{c}{FL+\\CARE(DSR)} & \tabincell{c}{FL+\\CARE(DeepLabV3)}\\
	\midrule
	\textbf{MCA} & 77.7 & 80.2 & \textbf{81.4} & 72.0    & 72.3  & \textbf{74.6} \\
	\tabincell{c}{Recall of the\\most minor class} & 56.5 & 78.3 & \textbf{86.7} & 52.2 & 60.9 & \textbf{82.6} \\
	\bottomrule
	\end{tabular}%
	\label{table:autobboxcombine}%
\end{table}%

%% file: tables/increase.tex
\begin{table*}[h]
		\centering
        \scriptsize
		\caption{Performance of CARE when masks were provided for different number of minority classes. In this table, CARE [X] means the masks were provided for the disease(s) X when the proposed CARE method was used for model training.}  
		\label{tab:more_minorities}
		\begin{tabular}{c|cccccccc}
		\toprule
		\multirow{2}{*}{Method} & \multirow{2}{*}{\textbf{MCA}} & \multicolumn{7}{c}{ \textbf{Recall}} \\
		\cline { 3 - 9 }  & & DF & VASC & AKIEC & BCC & BKL & MEL & NV \\
	\hline  CLS (Baseline) & $77.69$ & $56.52$ & $89.65$ & $68.18$ & $90.29$ & $75.45$ & $71.17$ & $92.54$ \\
	\hline  CARE [DF] & $\mathbf{8 1 . 3 5}$ &  \cellcolor{mypink}{$\mathbf{8 6 . 9 6}$} & $89.66$ & $66.67$ & $88.35$ & $76.82$ & $68.47$ & $92.54$ \\
			CARE [DF+VASC] & $80.51$ & \cellcolor{mypink}{$82.61$} & \cellcolor{mypink}{$\mathbf{9 6 . 5 5}$} & $66.67$ & $86.41$ & $74.09$ & $64.41$ & $9 2 . 8 4$ \\
			CARE [DF+VASC+AKIEC] & $80.28$ & \cellcolor{mypink}{$69.57$} & \cellcolor{mypink}{$89.66$} & \cellcolor{mypink}{$\mathbf{74.24}$} & $ 90.29 $ & $76.82$ & $68.92$ & $92.46$\\
	\bottomrule
	\end{tabular}
\end{table*}

%% file: sec/conclusion.tex
\section{Conclusions}\label{sec:conclusion}
In this study, we proposed a simple yet effective learning framework CARE to handle the data imbalance issue in medical image diagnosis. The attention mechanism is embedded into the CNN learning process for the minority category, helping the CNN focus on the right lesion region during learning and thus improving the classification accuracy for the minority class. CARE uses Grad-CAM to estimate attention maps and bounding boxes to indicate proper lesion regions of minority categories. This method can be applied to any CNN based classifier without altering neural network architectures. To alleviate the need for manual annotations of those bounding boxes, we further proposed variants of CARE with comparable performance. Comprehensive experiments have been performed on two commonly used imbalanced medical image datasets, showing that the proposed method can help the classifier improve the classification performance, particularly for the minority classes. Models trained together with CARE have a \textred{$2.5\text{-}4\%$} improvement on the Skin dataset and \textred{$6.4\text{-}8.3\%$} improvement on the Pneumonia dataset in mean class accuracy, and \textred{$4\text{-}21\%$} improvement on the smallest class of the Skin dataset and \textred{$25\text{-}38\%$} improvement on the smallest class of the Pneumonia dataset in the recall.
CARE can also combine with different existing data imbalance strategies to further boost their performances, demonstrating its flexibility.

The proposed CARE  method can be improved from the following aspects. First, {CARE} is a two-stage method in which the mask information is used only in the second stage to fine-tune the model.
In future work, we will investigate an end-to-end learning framework in which the mask information can be simultaneously estimated together with model training. \textred{Since there may be small lesions, we believe that replacing Grad-CAM with a more precise method for generating activation maps could improve the model's performance.}
 Second, CARE is mainly designed for CNN models. Considering that the recently proposed Capsule network~\cite{goceri2020capsnet,goceri2021analysis,goceri2021capsule} and vision Transformer~\cite{dosovitskiy2020image,liu2021swin} models can well capture spatial relationships of learned local features and have been successfully applied to image classification, it is interesting to extend the proposed CARE method to such model backbones in future work. \textred{While the proposed method showed promising results for the smallest class, applying CARE to multiple classes showed a decrease in recall for those classes with smaller sample sizes.} In addition, the CARE method is currently applied to two image classification tasks only. Its potential usage in more medical diagnosis applications and other medical image analysis tasks like lesion detection could be further investigated.

%% file: sec/appendix.tex
\section{Appendix}\label{sec:appendix}

\input{./algorithms/gradcam.tex}

\clearpage
\input{./algorithms/alg1.tex}

%% file: algorithms/gradcam.tex
\begin{algorithm}[H]
    \caption{\small Pseudocode of generation of feature maps in PyTorch-like style.}
    \label{alg:gradcam}
    \footnotesize
    \begin{algorithmic}[0]
        \STATE \textcolor{mygreen}{\# feature\_map: the activation feature map $A$ with the shape of [2048, 7, 7]}
    	\STATE \textcolor{mygreen}{\# grad\_class: the gradient score of class score $Y^c$ with respect to the activation feature map,   with the shape of [2048, 7, 7], which can be obtained by backpropagation.}
    	\STATE
    	\STATE def get\_gradCAM(feature\_map, grad\_class):
    	\STATE \quad\quad channels, height, width = grad.size()
    	\STATE \quad\quad grad = mean(grad, dim=[1, 2])
    	\STATE \quad\quad weight = grad.reshape(channels, 1, 1)
    	\STATE \quad\quad feature\_map = weight * feature\_map
    	\STATE \quad\quad grad\_cam = sum(feature\_map, dim=0)
    	\STATE \quad\quad grad\_cam = grad\_cam.reshape(7, 7)
    	\STATE \quad\quad return grad\_cam
    	\STATE
    	\STATE \textcolor{mygreen}{\# Normalize the feature map to the range of [0-1] and resize from [7, 7] to [224, 224].}
    	\STATE def post\_process(grad\_cam):
    	\STATE \quad\quad grad\_cam -= min(grad\_cam)
    	\STATE \quad\quad grad\_cam /= max(grad\_cam)
    	\STATE \quad\quad grad\_cam = interpolate(grad\_cam, size=(224, 224), mode='bilinear')
    	\STATE \quad\quad return grad\_cam
    \end{algorithmic}
\end{algorithm}

%% file: algorithms/alg1.tex
\begin{algorithm}[H]
    \caption{\small Pseudocode of CARE in PyTorch-like style.}
    \label{alg:1}
    \footnotesize
    \begin{algorithmic}[0]
        \STATE \textcolor{mygreen}{\# loader\_m: dataloader with masks for the small class}
        \STATE \textcolor{mygreen}{\# loader: dataloader without masks for the small class}
    	\STATE \textcolor{mygreen}{\# m\_p, m\_f: model respectively for the pretrain stage and the (attention based) finetune stage}
    	\STATE \textcolor{mygreen}{\# $\alpha$, $\lambda$, $\tau$: hyperparameters}
    	\STATE
        \STATE \textcolor{mygreen}{\# the pretrain stage}
        \STATE for epoch in range(n\_epochs):
    	\STATE \quad\quad for x, labels in loader: \textcolor{mygreen}{\# load a minibatch x and labels with N samples}
    	\STATE \quad\quad\quad\quad x = aug(x)
    	\STATE \quad\quad\quad\quad logits = m\_p.forward(x)
        \STATE \quad\quad\quad\quad \textcolor{mygreen}{\# pre-train with weighted cross-entropy loss}
    	\STATE \quad\quad\quad\quad loss = CrossEntropyLoss(logits, labels) 
    	\STATE \quad\quad\quad\quad loss.backward()
    	\STATE \quad\quad\quad\quad update(m\_p)
    	\STATE 
        \STATE \textcolor{mygreen}{\# the (attention based) finetune stage}
    	\STATE m\_f.params = m\_p.params \textcolor{mygreen}{\# initialize with parameters from the pretrain stage}
    	\STATE m\_f.register\_hook() \textcolor{mygreen}{\# Register function to store feature map and gradient score.}
        \STATE for epoch in range(n\_epochs):
    	\STATE \quad\quad \textcolor{mygreen}{\# load a minibatch x, Mask M and labels. Masks are only for the small class}
    	\STATE \quad\quad for x, M, labels in loader\_m: 
    	\STATE \quad\quad \quad\quad x, M = aug(x, M) \textcolor{mygreen}{\# augmented with the same operation}
    	\STATE \quad\quad \quad\quad logits, feature\_map = m\_f.forward(x) 
        \STATE \quad\quad\quad\quad \textcolor{mygreen}{\# finetune with weighted cross-entropy loss}
		\STATE \quad\quad\quad\quad $L_c$ = (1-$\alpha$) * CrossEntropyLoss(logits, labels)		
        \STATE \quad\quad\quad\quad \textcolor{mygreen}{\# generate feature map F by Grad-CAM style}
		\STATE \quad\quad\quad\quad \textred{$L_c$.backward(retrain\_graph=True)} 
        \STATE \quad\quad\quad\quad grad\_class = m\_f.get\_gradients() \textcolor{mygreen}{\# gradient score with respect to feature map}
        \STATE \quad\quad\quad\quad F = get\_gradCAM(feature\_map, grad\_class) \textcolor{mygreen}{\# obtain Grad\_CAM, and defined in Algorithm 1}
        \STATE \quad\quad\quad\quad F = post\_process(F) \textcolor{mygreen}{\# Defined in Algorithm 1}
		\STATE \quad\quad\quad\quad if M is not empty:
		\STATE \quad\quad\quad\quad\quad\quad $L_{in}$ computed by Equation (3) in \textcolor{blue}{Section 3.1}
		\STATE \quad\quad\quad\quad\quad\quad $L_{out}$ computed by Equation (4) in \textcolor{blue}{Section 3.1}
		\STATE \quad\quad\quad\quad\quad\quad $L_a=L_{in}+\lambda L_{out}$
		\STATE \quad\quad\quad\quad else:
		\STATE \quad\quad\quad\quad\quad\quad $L_a=0$ 
		\STATE \quad\quad\quad\quad $L_a *= \alpha$
		\STATE \quad\quad\quad\quad $L_a$.backward()
		\STATE \quad\quad\quad\quad update(m\_f)
    \end{algorithmic}
\end{algorithm}